\documentclass{article} 
\usepackage{iclr2027_conference,times}

\usepackage{amsmath,amsfonts,bm}

\def\eqref#1{equation~\ref{#1}}

\def\1{\bm{1}}

\DeclareMathAlphabet{\mathsfit}{\encodingdefault}{\sfdefault}{m}{sl}
\SetMathAlphabet{\mathsfit}{bold}{\encodingdefault}{\sfdefault}{bx}{n}

\usepackage{hyperref}
\usepackage{url}
\usepackage{booktabs}
\usepackage{amsmath}
\usepackage{graphicx}

\iclrfinalcopy

\title{A Lie Detector Test for Language Models:\\ Reading Knowledge a Model Won't Reveal}

\author{Hiskias Dingeto \\ StackOne Technologies \\ \texttt{hiskias@stackone.com}}

\begin{document}
\maketitle
\lhead{Preprint}

\begin{abstract}
Large language models can hold knowledge they do not report. A model may sandbag on a
capability evaluation, or answer against what it internally knows, and its outputs alone
cannot tell whether it is hiding an answer or simply does not have one. We borrow the
Concealed Information Test, a forensic method that identifies guilty knowledge by presenting
a suspect with the true detail among plausible decoys and measuring a stronger response to
the item they recognize. Our method, Probe of Internal Recognition (PIR), does the same
inside a model. It presents a question with its candidate answers and reads, from the
model's internal states, which candidate the model recognizes as correct. PIR is
reference-free, needing no honest reference model and no labeled truth corpus. Across eight models from five families (Gemma, Qwen, Llama, Mistral, and Phi), PIR recovers the
recognized answer at 0.70 to 0.87 balanced accuracy, well above the 0.28 to 0.40 unknown-item baseline and
the 0.25 chance rate. It stays readable across every form of concealment we test, from prompted
deception and trained sandbagging to external password-locked and circuit-broken checkpoints, with
recognition between 0.85 and 0.93. When the model hides a known answer, recognition
stays high. When unlearning removes the knowledge, recognition drops to the level of a question the
model never knew. PIR
therefore separates a model that will not answer from one that cannot, which supports
sandbagging audits and unlearning verification. The signal is causal, adds information
beyond black-box behavioral cues, and extends from multiple-choice questions to free-form
generation.
\end{abstract}

\section{Introduction}

Capability evaluations assume a model tries to answer. This assumption can fail. A model
can underperform on purpose, a behavior called sandbagging, or otherwise return an answer
that conflicts with what it internally represents \citep{vanderweij2024sandbagging}. This
matters for safety. A safety case that rests on a dangerous-capability evaluation is only
trustworthy if the model is not hiding what it can do. Unlearning raises a similar concern.
When a developer removes hazardous knowledge, we want to confirm the knowledge is gone and
not merely hidden at the output \citep{li2024wmdp}.

Behavior alone cannot resolve these questions. A model that hides an answer and a model
that never knew it can produce the same wrong output. To tell them apart we have to look
inside the model.

A line of work shows this is feasible. The truth of a statement is often encoded linearly
in a model's activations, and simple probes recover it even when the model outputs something
false \citep{orgad2024llms, marks2023geometry, azaria2023internal}. These results leave open
questions that matter for safety. The probes are usually supervised or need a labeled truth
corpus, and some designs assume a separately trained honest model. They are shown mainly on the
truthfulness of declarative statements. To our knowledge, they have not been evaluated as a
test that separates concealment from genuine removal, nor against models trained to conceal
or against unlearning. This leaves a concrete question. Can we read a model's concealed knowledge
without a truth reference, and can one readout tell hiding apart from erasure, in the
settings safety cares about?

We answer both by borrowing a tool from deception detection. The Concealed Information Test
identifies guilty knowledge without a confession. It presents a suspect with the true detail
among plausible decoys and looks for a stronger response to the item the suspect recognizes
\citep{lykken1959gsr, verschuere2011memory, benshakhar2003validity}. Our method, Probe of
Internal Recognition (PIR), runs the same test inside a model, reading a recognition signal from
activations much as neural memory detection reads it from a brain response \citep{farwell1991truth}. Given a question and a set of candidate answers, PIR reads from the
model's activations which candidate it recognizes as correct, using the contrast between the
correct option and its distractors. PIR is reference-free. It needs no honest reference model and
no labeled truth corpus, because the distractors supply the contrast.

PIR recovers concealed knowledge across models and forms of concealment. Under prompted
concealment the readout reproduces across the Gemma, Qwen, Llama, Mistral, and Phi families, and
it recovers answers from models trained to sandbag and from external password-locked and
circuit-broken checkpoints we did not build \citep{hofstaetter2025elicitation, zou2024circuit}.
It then marks a clear split between hiding and erasure. When a model hides an answer it knows, the
answer stays decodable, and a divergence detector calibrated on the model's own base checkpoint
flags the concealment. When unlearning removes the knowledge, recognition falls to the unknown-item baseline, both for our own run and for external unlearning checkpoints across families
\citep{li2024wmdp, zhang2024npo}. The same readout therefore separates a model that will not
answer from one that cannot, which is what sandbagging audits and unlearning verification need.
The recognition direction is used by the model, since steering along it changes the answer while a
matched output-direction control does not. The readout also adds information beyond black-box
behavioral cues, most when a model conceals confidently, and a label-free detector built from the
same signal flags concealed knowledge in free-form generation. This holds for models that are not
trained to evade it, and a model adversarially trained to obfuscate its recognition can defeat the
readout.

We make the following contributions.
\begin{itemize}
\item \textbf{PIR}, a reference-free readout that recovers a model's concealed answer from
its internal recognition of the correct option among distractors, adapted from the Concealed
Information Test.
\item \textbf{A hiding-versus-erasure boundary.} Concealment that keeps the model's capability
recoverable leaves the readout intact, while unlearning that removes the capability erases it.
This gives a practical signal for sandbagging audits and unlearning verification.
\item \textbf{Evidence that the signal is used and additive.} Steering along the recognition
direction changes the model's answer, and the readout adds information over black-box
behavioral detection.
\item \textbf{Generalization to free-form generation}, where candidates are reconstructed
from the model's own samples instead of supplied by the task.
\end{itemize}

\section{Related Work}

\paragraph{Reading knowledge and truth from internal states.} A model's internal states often
encode whether a statement is true, and linear probes recover this signal even when the model
outputs something false \citep{azaria2023internal, marks2023geometry, orgad2024llms,
burger2024truth}. \citet{orgad2024llms} show that a model can internally represent the correct
answer while generating a wrong one, and \citet{mallen2024quirky} read the correct answer, the
model's latent knowledge, from models finetuned to make context-dependent errors, an early
analogue of sandbagging. Some methods aim to be reference-free. \citet{burns2023ccs} recover
latent knowledge from unlabeled activations
through logical consistency, though later work questions whether such objectives find knowledge
or the most salient feature \citep{farquhar2023challenges}. PIR builds on this line and departs
from it. It reads knowledge reference-free through a Concealed Information Test contrast between
the correct option and its distractors, so it needs no truth label and no consistency assumption. It
separates hiding from erasure. And it extends beyond declarative statements to free-form
generation.

\paragraph{Probing, steering, and representation engineering.} Reading a concept from
activations and steering it are two sides of one method. Linear classifier probes date back to
\citet{alain2016understanding}. Representation engineering reads and controls high-level concepts
such as honesty \citep{zou2023repe}, and activation steering adds concept directions to the
residual stream to change behavior \citep{turner2023activation, rimsky2024steering}. Inference-time
intervention shifts activations along a truthfulness direction to elicit truthful answers
\citep{li2023iti}. \citet{campbell2023localizing} localize and causally flip instructed lying. Our recognition
direction uses the same contrast-pair construction, and our causal test steers it to change the
model's answer, with a matched control that steers the output direction and does not.

\paragraph{Detecting deception and sandbagging.} A growing body of work detects deception and
hidden capability. Black-box methods ask unrelated follow-up questions to flag lies
\citep{pacchiardi2023liar}, and benchmarks separate honesty from accuracy \citep{ren2025mask}.
Behavioral lie detectors that make a model self-report or classify its own deception generalize
poorly across deception types and cannot recover a fact the model does not report
\citep{hopkins2026liedetectors}, a limit PIR avoids by reading internal recognition directly.
White-box probes detect deceptive responses and backdoored behavior at high accuracy
\citep{goldowskydill2025deception, macdiarmid2024probes}, and frontier models can scheme and
sandbag on evaluations \citep{meinke2024scheming}, and trained deceptive behavior can persist
through safety training \citep{hubinger2024sleeper}. The sandbagging threat
is studied through models that underperform on purpose \citep{vanderweij2024sandbagging} and
password-locked organisms whose capability must be elicited \citep{greenblatt2024passwordlocked,
hofstaetter2025elicitation}. \citet{tice2024noise} elicit hidden capability by injecting weight
noise, a reference-free detector like ours. PIR recovers the concealed answer itself, works
reference-free, and adds information over black-box behavioral cues. Where prior probes output a
deception flag, PIR outputs which answer the model recognizes. We evaluate it on external
password-locked and circuit-broken organisms \citep{hofstaetter2025elicitation}.

\paragraph{Does unlearning erase or suppress?} Machine unlearning removes hazardous knowledge
from a model. RMU (Representation Misdirection for Unlearning) misdirects the representations of
forgotten content \citep{li2024wmdp}, and NPO (Negative Preference Optimization) reframes
forgetting as preference optimization to improve on gradient-ascent forgetting
\citep{zhang2024npo, yao2023llmunlearning}. A recent line argues that these methods
suppress knowledge instead of removing it, since light finetuning, benign relearning, or small
weight edits recover much of the forgotten capability \citep{lucki2024adversarial, hu2024jogging,
deeb2024doremove, che2025modeltampering}, which motivates evaluation beyond input-output behavior
\citep{lynch2024eight}. Our results add a different view. The recognition signal that
PIR reads is driven to the unknown-item baseline by unlearning, most completely by RMU, on the
unlearned domain, even though the same signal survives every concealment that keeps the
capability recoverable, including prompt deception, trained sandbagging, and circuit-breaking.
We report this as a
property of the readout at deployment, and we do not claim the weights can never be made to
relearn.

\paragraph{Robustness of probes to adversarial evasion.} Fixed activation probes can be evaded by
an adversary with white-box access. \citet{bailey2025obfuscated} optimize a model's activations to
bypass latent-space defenses, driving probe recall to zero while keeping the harmful behavior, and
probe-based deception detectors carry the same caveat \citep{goldowskydill2025deception}. PIR
inherits this limit. It is an audit signal for models that are not trained to evade it, and a
model adversarially trained to obfuscate its recognition can defeat it.

\section{Method}
\label{sec:method}

\begin{figure}[t]
\centering
\includegraphics[width=\linewidth]{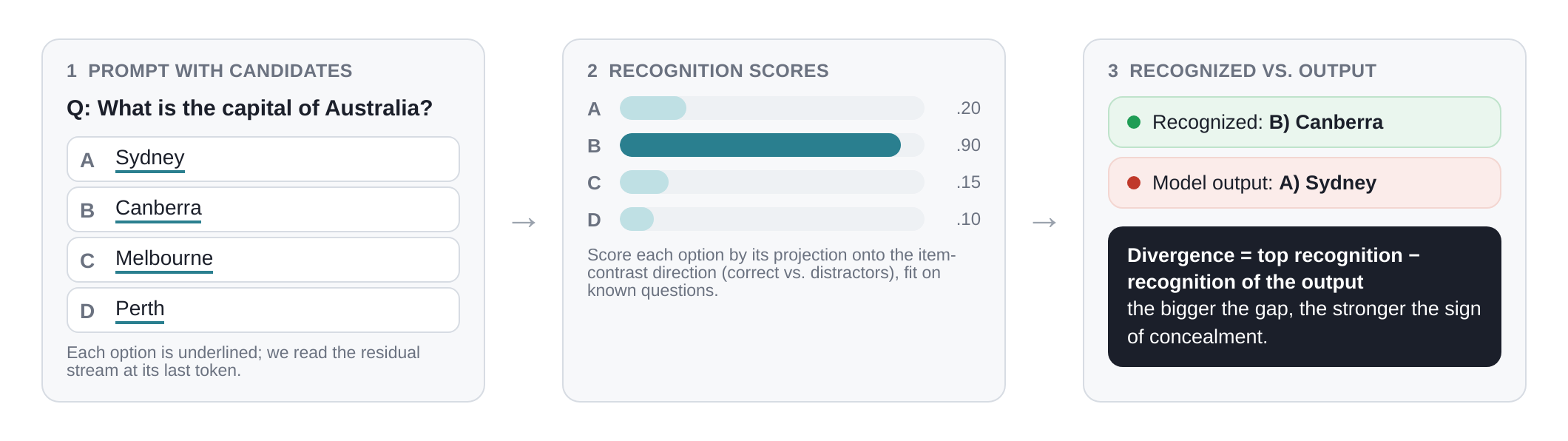}
\caption{The PIR pipeline. We read each option's last-token span, project onto the item-contrast
direction, and take the highest-scoring option as the recognized answer. The divergence detector
compares that answer against the model's output.}
\label{fig:pipeline}
\end{figure}

We describe PIR in three parts. We read the answer the model recognizes, we turn that readout into a
label-free concealment flag, and we use the flag to separate concealment from erasure.
Figure~\ref{fig:pipeline} gives the pipeline.

\subsection{Setup and notation}
We study multiple-choice questions. A question $q$ has $k_q$ candidate options $o_1,\dots,o_{k_q}$, one
of which, at index $c(q)$, is correct. We present $q$ with its options to a model $M$ and read the
residual stream. For option $o_j$ we take the hidden state at its last token, which we call the option
\emph{span}, and write $x_{q,j}^{\ell}\in\mathbb{R}^{d}$ for that vector at layer $\ell$. The last token
is the strongest read position, ahead of a mean over the option's tokens and well ahead of the first
token (Appendix~\ref{app:ablations}). We use three conditions of a
model. In the \emph{honest} condition the model answers with no pressure to conceal. In the
\emph{eliciting} condition we recover its capability with a prompt, password, or trigger. In the
\emph{deployed} condition the model may conceal. The \emph{base model} is the model's own checkpoint
before sandbagging or unlearning. We say $M$ \emph{knows} $q$ if it answers correctly in the honest
condition, and $M$ \emph{conceals} on $q$ if it knows $q$ yet outputs an index $\hat{y}\neq c(q)$. The
\emph{unknown-item baseline} is the recognition level on questions the model does not know, which serves as
our reference level. Our goal is to recover $c(q)$ from the activations $\{x_{q,j}^{\ell}\}$ when
$\hat{y}$ is wrong. Figure~\ref{fig:pipeline} summarizes the readout.

\subsection{Reading the answer a model recognizes}
PIR reads which option $M$ recognizes as correct by contrasting the correct option against its distractors.
On a set $\mathcal{K}$ of the model's own known questions, we form the \emph{item-contrast direction}
\begin{equation}
w^{\ell} \;=\; \frac{1}{|\mathcal{K}|}\sum_{q\in\mathcal{K}} \Big( x_{q,c(q)}^{\ell} - \frac{1}{k_q-1}\sum_{j\neq c(q)} x_{q,j}^{\ell} \Big),
\end{equation}
the mean difference between the correct-option span and the average distractor span. For a new question we
score each option by its projection onto this direction and decode the recognized answer as the
highest-scoring option,
\begin{equation}
r_j \;=\; \langle x_{q,j}^{\ell}, w^{\ell}\rangle, \qquad \hat{c} \;=\; \arg\max_j r_j .
\end{equation}
We choose the layer $\ell$ by question-disjoint cross-validation, so the direction is never fit and
evaluated on the same question. We call $w^{\ell}$ the recognition direction, and use that name for it
in the rest of the paper.

This readout is \emph{reference-free}. It needs no separately trained honest model and no external
truth corpus, because the distractors supply the contrast and the labels for $\mathcal{K}$ are the model's
own correct answers in the honest condition. When no labels are available, the top principal component
of the per-question centered option spans recovers a similar direction, so the readout can also run
without labels. Because the contrast is taken within a question, between an option and its distractors, it
reads the option the model treats as correct, which addresses the concern that unsupervised objectives
may track the most salient feature of the text \citep{farquhar2023challenges}. On models whose
activations carry a few very large dimensions, we standardize the spans before the projection;
Appendix~\ref{app:norm} gives details.

\subsection{Detecting concealment without labels}
At deployment we do not have the correct answer, so we turn the readout into a label-free flag. The
\emph{divergence detector} compares the recognized answer against the model's output. It scores each
option with a direction $\tilde{w}^{\ell}$ calibrated \emph{off-distribution}, fit on the base model
answering the same questions in the honest condition and then applied to the deployed model's spans,
so the direction is never fit on the concealed activations it scores. Writing
$\tilde{r}_j = \langle x_{q,j}^{\ell}, \tilde{w}^{\ell}\rangle$, the detector score for a question with
output $\hat{y}$ is
\begin{equation}
D \;=\; \max_j \tilde{r}_j \;-\; \tilde{r}_{\hat{y}},
\end{equation}
the recognition score of the top-recognized option minus that of the option the model output. $D$ is
small when the model outputs what it recognizes, or when no option is strongly recognized, and large
when the model recognizes one option yet outputs another.

The base model here is the model's own pre-modification checkpoint, so the detector needs that
checkpoint. This is available for unlearning verification and for auditing a released model against a
tampered copy. It does not require a separately trained model that must be trusted as honest. We
evaluate the detector by how well it separates two groups of questions. A question is \emph{known} if
the base model answers it correctly in the honest condition across all $k_q$ cyclic rotations of the
option order, a strict test that removes position bias and lucky guesses. A deployed instance is \emph{conceal} if the question is
known and the deployed output is wrong, and \emph{unknown} if the question is not known. From the
output alone these two look the same, since both give a wrong answer. We report the detector as the
area under the ROC curve for separating conceal from unknown.

\subsection{Distinguishing concealment from erasure}
\begin{figure}[t]
\centering
\includegraphics[width=\linewidth]{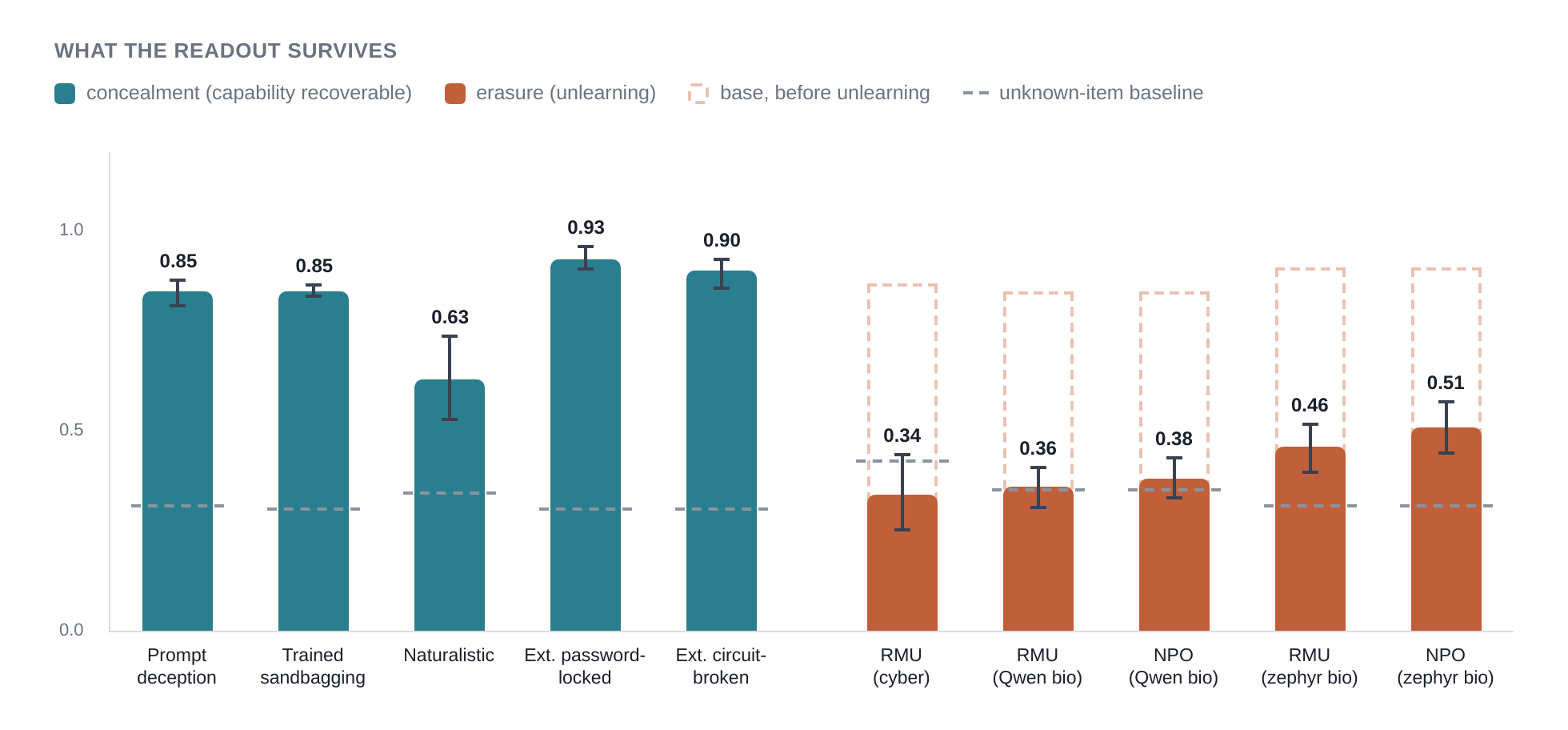}
\caption{What the readout survives. Gray ticks mark the per-condition unknown-item baseline, dashed
outlines mark base recognition before unlearning, and whiskers are bootstrap 95\% confidence intervals.}
\label{fig:boundary}
\end{figure}
The readout also tells whether a wrong output hides an answer the model holds or reflects an answer
that has been removed. We separate two regimes by whether the capability is recoverable at inference
time, meaning by a prompt, a password, or a trigger, with no weight update. Under \emph{concealment}
the capability remains and inference-time elicitation recovers it. Prompt deception, a trained
sandbagging policy, and circuit-breaking all fall here, the knowledge is present, and PIR reads it.
Under \emph{erasure} inference-time elicitation does not recover the capability. Unlearning methods
such as RMU and NPO fall here \citep{li2024wmdp, zhang2024npo}, and the readout falls to the unknown-item baseline. This inference-time criterion is what places unlearning on the erasure side. It does not
require the knowledge to be unrecoverable by every means, and light finetuning can relearn some of it
\citep{lucki2024adversarial}, which we do not dispute. Because the criterion is a behavioral test that
does not use PIR, the boundary is a property we verify, and it does not define the readout in terms of
itself.

\subsection{Extending to free-form generation}
Free-form questions provide no option list, so we reconstruct one. We sample answers from the model
under both an eliciting and a deployed condition and take the distinct normalized strings as
candidates. Each candidate is placed in a short completion, ``Question: $q$ Answer: $o_j$'', and we
read its last-token span, as in the multiple-choice case. The item-contrast readout then decodes which
candidate the model recognizes as correct. In free-form the conceal and unknown groups are hard to
separate by the divergence score, because the reconstructed candidate set always offers some recognized
option. We instead flag concealment by the \emph{peak recognition} $\max_j r_j$, which is high when the
model strongly recognizes a candidate and low when it recognizes none. This peak-recognition detector
flags concealed knowledge in free-form for a trained sandbagging policy as well as a prompted one.

\section{Experiments}
We test PIR across models and forms of concealment, at the hiding-versus-erasure boundary, at
deployment, and in free-form generation.

\subsection{Setup}
\label{sec:setup}
\begin{figure}[t]
\centering
\includegraphics[width=\linewidth]{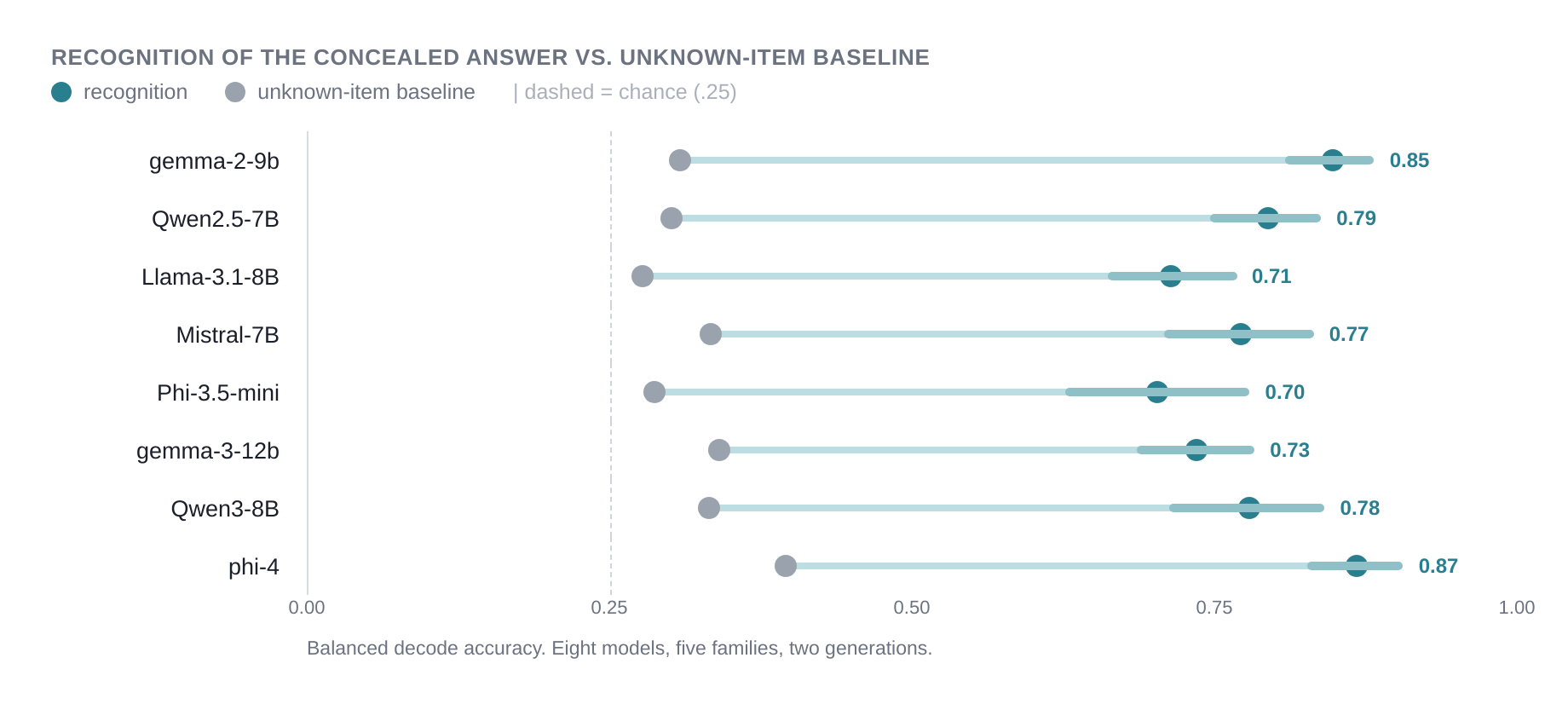}
\caption{Recognition of the concealed answer against the unknown-item baseline under prompt deception,
for eight models across five families and two generations. Whiskers are bootstrap 95\% confidence
intervals.}
\label{fig:crossmodel}
\end{figure}
We use \texttt{gemma-2-9b} (the instruction-tuned checkpoint) as the primary model. It is a well
studied mid-size instruction model, and both the external sandbagging organisms and the external
unlearning checkpoints we test are built on comparable open models, which lets us reuse one pipeline.
We also reproduce the readout on eight models from five families (Gemma, Qwen, Llama, Mistral, Phi)
across two generations, including \texttt{phi-4}, on which we also confirm the full pipeline. Datasets
are MMLU, the three WMDP (Weapons of Mass Destruction Proxy) domains (biology, chemistry,
cybersecurity), ARC-Challenge, and TriviaQA for free-form. Concealment comes in several forms.
\emph{Prompt deception} instructs a model to answer against what it knows. A \emph{trained sandbagging
policy} is a LoRA organism that answers correctly under a trigger and wrongly without it, in a
prefix-triggered and a naturalistic (imitate-weak) variant. We also test two external organisms we did not build, a
password-locked and a circuit-broken \texttt{gemma-2-9b} \citep{hofstaetter2025elicitation}. Erasure
comes from unlearning, RMU and NPO, both our own runs and external checkpoints on Qwen2.5-7B and
zephyr-7b-beta \citep{li2024wmdp, zhang2024npo}. We report recognition as balanced decode accuracy over
the options, with the unknown-item baseline as the reference level, and detectors as AUC on the
conceal-vs-unknown split. All readouts use question-disjoint cross-validation and bootstrap 95\%
confidence intervals. The raw readout is fragile on models with a few very large activation dimensions,
so we standardize before the projection (Appendix~\ref{app:norm}); on Gemma-3 this moves recognition
from $0.389$ to $0.735$.

\subsection{Reading concealed knowledge across models}
Under prompt deception, PIR recovers the concealed answer well above the unknown-item baseline on every
model. Across eight models from five families and two generations, recognition ranges from $0.70$ to
$0.87$ while the unknown-item baseline stays between $0.28$ and $0.40$ (near $0.30$ for most models), a gap
of $+0.40$ to $+0.54$ (Figure~\ref{fig:crossmodel}). A control that reads the model's output option is
high everywhere, and no model recovers the wrong output as the recognized answer, so the readout tracks
the concealed answer. A belief-vs-output test confirms this. A binary within-condition probe that scores
$0.985$ by reading the output drops to chance under the test, while the item-contrast readout survives
it (Appendix~\ref{app:stats}).

The recognition direction is causal. On held-out known questions we add the direction, scaled by a dose
$\alpha$, to the residual stream at the tokens of a wrong option,
\begin{equation}
x_{q,j}^{\ell} \;\leftarrow\; x_{q,j}^{\ell} + \alpha\, w^{\ell},
\end{equation}
and check whether the model then outputs that option. This makes \texttt{gemma-2-9b} pick the steered
option with probability $0.189$ (Wilson CI $[0.126, 0.274]$), against $0.005$ for 150 matched random
directions (permutation $p<1/150$) and monotone in $\alpha$. A matched control that adds the option
letter's output direction at the same tokens
does nothing ($0.000$), and the recognition direction is nearly orthogonal to it (cosine $0.01$). The
steer reproduces on \texttt{phi-4} ($0.122$, CI $[0.074, 0.194]$). The direction is used by the model
upstream of the output, and it does something the output direction does not.

\subsection{What the readout survives}
\begin{table}[t]
\centering
\caption{Detector AUC (conceal vs.\ unknown). Internal is the off-distribution recognition detector;
combined adds the black-box features.}
\label{tab:detector}
\begin{tabular}{lcccc}
\toprule
Organism & internal & text-only & black-box & combined \\
\midrule
trained prefix (\texttt{gemma-2}) & 0.718 & 0.517 & 0.645 & -- \\
external password-locked & 0.892 & 0.660 & 0.845 & \textbf{0.929} \\
external circuit-broken & 0.848 & 0.621 & 0.860 & \textbf{0.899} \\
\bottomrule
\end{tabular}
\end{table}
The readout stays high under output-level concealment and collapses under erasure
(Figure~\ref{fig:boundary}). Under a trained sandbagging policy on MMLU, recognition on concealed
questions is $0.85$ (CI $[0.83, 0.87]$) against a unknown-item baseline of $0.30$, and the naturalistic
organism gives $0.63$ (CI $[0.52, 0.74]$) on the concealed subset and $0.81$ on the full known set
against $0.34$. On the two external organisms, recognition is $0.93$ (CI $[0.90, 0.97]$, password-locked)
and $0.90$ (CI $[0.85, 0.93]$, circuit-broken) against the unknown-item baseline near $0.30$, so PIR reads a sandbagger
built by others and reads through circuit-breaking.

Erasure is different. On a strict-gated run, unlearning \texttt{gemma-2-9b} with RMU drops base
recognition of $0.85$ (CI $[0.78, 0.92]$) to $0.39$ (CI $[0.30, 0.49]$), which overlaps the unknown-item baseline at $0.35$ (CI $[0.28, 0.41]$). The drop is domain-specific. A run that unlearns only
cybersecurity drives cyber recognition to $0.34$ (CI $[0.25, 0.45]$), below its unknown-item baseline of
$0.42$, while biology ($0.91$ to $0.68$, CI $[0.62, 0.75]$) and chemistry ($0.76$ to $0.51$, CI
$[0.42, 0.59]$) stay readable. The erasure replicates on external checkpoints across families. On
Qwen2.5-7B, biology recognition falls from $0.85$ to $0.36$ (RMU, CI $[0.31, 0.41]$) and $0.38$ (NPO, CI
$[0.33, 0.44]$), at the unknown-item baseline of $0.35$; on zephyr-7b-beta from $0.91$ to $0.46$ (RMU, CI
$[0.39, 0.52]$) and $0.51$ (NPO, CI $[0.44, 0.58]$, a partial drop). Domain-specificity is clean on our
own runs, and the external checkpoints over-forget, so we report them as a cross-model erasure
replication.

\subsection{Detecting concealment at deployment}
At deployment the divergence detector, calibrated off-distribution, separates a model that hides an
answer from one that does not know it. On the trained \texttt{gemma-2-9b} organism, the off-distribution
direction decodes the concealed answer at $0.82$, against $0.21$ for the off-distribution unknown-item baseline, which can fall below chance, and $0.30$ for a text-only decode, so the signal generalizes
off-distribution and is not an option-text prior. As a
detector it reaches AUC $0.718$ (CI $[0.695, 0.739]$) on the hard conceal-vs-unknown split, above a
text-only baseline at $0.517$ ($+0.20$, CI $[0.17, 0.23]$) and above a combined black-box behavioral
detector at $0.645$ ($+0.073$, CI $[0.048, 0.097]$).

Whether internals beat black-box depends on how the organism conceals (Table~\ref{tab:detector}). In
that table the internal detector is single-layer for the prefix organism on MMLU and multilayer for the
external organisms on WMDP, so we compare within each row. The
two external organisms sandbag by emitting a low-confidence wrong answer, so the concealment leaks into
the output distribution and black-box is strong. Once compute-matched and optimized, the internal
readout beats black-box on the password-locked model ($0.89$ vs $0.85$) and ties it on the
circuit-broken one ($0.85$ vs $0.86$). On both, the internal signal is additive. The combined monitor
is best, at $0.93$ and $0.90$, with the internal contribution significant over black-box alone. So
internals add information the outputs do not carry, and they dominate when a model conceals confidently.
As concealment becomes confident, the concealed output looks like an honest answer and black-box
detection falls toward chance, while the internal readout holds (Appendix~\ref{app:frontier}).

\subsection{Beyond multiple choice}
PIR extends to free-form generation, where the task gives no option list. We reconstruct candidates from
the model's own samples on TriviaQA. Free-form has no clean unknown set, so here we compare recognition
to chance in place of the unknown-item baseline. On the honest model, restricted to questions where it produces both
a correct and a wrong candidate, the readout decodes the correct answer at $0.71$, above chance at
$0.42$, above a text-only decode at $0.54$, and above self-consistency at $0.56$
(Table~\ref{tab:freeform}). The readout reads knowledge, and it reads it better than the model's own
output distribution. Under a trained free-form sandbagging policy the reading survives. Recognition
decodes the concealed answer at $0.74$ against a text-only decode of $0.68$, and the peak-recognition
detector separates conceal from unknown at AUC $0.776$, above a behavioral baseline ($+0.123$, CI
$[0.052, 0.195]$). The decode advantage over text is small on factual questions, where the gold answer
is also the most relevant candidate, so we report the detector as the free-form claim.

\begin{table}[t]
\centering
\caption{Free-form results on TriviaQA. Decode is recognition decode accuracy; the detector is
peak-recognition AUC (conceal vs.\ unknown). Dashes mark not-applicable metrics.}
\label{tab:freeform}
\begin{tabular}{lcccc}
\toprule
Setting & decode & text-only & self-consistency & detector \\
\midrule
honest model & 0.71 & 0.54 & 0.56 & -- \\
trained sandbagger & 0.74 & 0.68 & -- & 0.776 \\
\bottomrule
\end{tabular}
\end{table}

\subsection{Evading the readout}
The readout is an audit signal for models that are not trained to evade it. We train an anti-probe
organism that keeps its capability under the trigger, sandbags without it, and is optimized to drive the
recognition margin to zero. Under question-disjoint cross-validation this organism drives both a
deployment-refit monitor and a fixed base-calibrated monitor down to the unknown-item baseline
while keeping capability ($0.71$). An earlier report that a fixed direction reads through the attack was
an in-sample artifact and is retracted. This matches the general finding that a white-box adversary can
obfuscate activations against a fixed probe \citep{bailey2025obfuscated}. The attack leaves an
option-span collapse that may be a detectable tamper, but that defense needs a base-model anomaly
reference and an anomaly-aware adversary, which we leave open. Appendix~\ref{app:adversarial} gives the
training and the dose sweep.

\section{Discussion and Limitations}
A reference-free item contrast reads the answer a model recognizes as correct even when its output
differs. This extends the finding that truth is linearly present in activations
\citep{orgad2024llms, marks2023geometry}; our contribution is what we do with it. The boundary is
the most useful part for safety, and two audits
follow. For sandbagging, a divergence between the recognized answer and the output flags a model
hiding capability on an evaluation. For unlearning, a readout that falls to the unknown-item baseline is
evidence that the knowledge is gone and not held back at the output. This second audit is also a
caution. Because circuit-breaking leaves the readout intact, a model that looks unlearned to
elicitation can still carry the knowledge, and PIR sees it.

We separate what the evidence shows from what it does not. The steering result shows the recognition
direction is used by the model, since adding it changes the answer and a matched output-direction
control does not. We do not claim the direction is necessary, and a variance-matched ablation did not
establish necessity. The detector adds information over black-box cues and dominates when a model
conceals confidently (Appendix~\ref{app:frontier}), while a low-confidence wrong answer already leaks to black-box detection, where
internals mainly add to it.

PIR also has clear limits. It needs a set of candidate answers, so it covers multiple-choice and,
through reconstructed candidates, free-form generation, but not open-ended reasoning with no enumerable
answer. The deployment detector
needs the model's own base or honest checkpoint, available for unlearning verification and release
auditing but not for a fully black-box system. A model trained to evade the readout can defeat it, so
the safety claim holds for models that are not adversarial to the probe. PIR reads only intact
representations, so it cannot tell erased knowledge from knowledge never had, since both give a silent
readout. The conceal and unknown labels come from a strict rotation-consistency gate, which discards
questions the model does not answer robustly and shapes the detector numbers. The clean domain-specific
erasure result rests on our own single RMU run, because the external checkpoints over-forget. Several
organisms are our own and single-seed, external validation does not cover every concealment type, and
the free-form detector is shown on one model and on factual questions.

\section{Conclusion}
A wrong answer from a model is ambiguous. It can hide knowledge the model holds or reflect knowledge it
never had. We adapt the Concealed Information Test to model activations, reading the concealed answer with
a reference-free item contrast that tells hiding apart from genuine absence, the distinction sandbagging
audits and unlearning verification need. The audit holds for models that are not trained to evade it.
As safety cases come to rest on capability evaluations and on unlearning, a reference-free read of what a
model recognizes gives an auditor a signal its outputs cannot provide. We hope the boundary between
concealment and erasure, and the means to test it, prove useful beyond the settings we study here.

\section*{AI use statement}
We used generative AI to draft and polish the text, to find and check related work, and to help
design experiments and write code. We did not use it to generate synthetic datasets or prove
mathematical claims. The authors verified every claim, number, citation, and code output, and take
responsibility for the final content.

\section*{Ethics statement}
This work is a defensive audit method. Our experiments use public model organisms and the public WMDP
benchmark, and we release no new hazardous capability. The most sensitive artifact is the anti-probe
organism, which shows that a model can be trained to evade the readout. We report it so that developers
and evaluators can anticipate the attack, and it confers no capability beyond defeating this monitor. We
recommend using the readout as one signal within an audit and pairing it with other checks.

\section*{Reproducibility statement}
Section~\ref{sec:method} defines the readout and the divergence detector, Section~\ref{sec:setup} lists
the models, datasets, and organisms, and the appendix gives the standardization and statistics. Experiments ran on single A100
GPUs. We will release the code and analysis scripts; the external organisms and unlearning checkpoints
we use are already public.

\bibliography{references}

@inproceedings{orgad2024llms,
  title={{LLMs} Know More Than They Show: On the Intrinsic Representation of {LLM} Hallucinations},
  author={Orgad, Hadas and Toker, Michael and Gekhman, Zorik and Reichart, Roi and Szpektor, Idan and Kotek, Hadas and Belinkov, Yonatan},
  booktitle={International Conference on Learning Representations (ICLR)}, year={2025}, note={arXiv:2410.02707}
}

@inproceedings{marks2023geometry,
  title={The Geometry of Truth: Emergent Linear Structure in Large Language Model Representations of True/False Datasets},
  author={Marks, Samuel and Tegmark, Max},
  booktitle={Conference on Language Modeling (COLM)}, year={2024}, note={arXiv:2310.06824}
}

@inproceedings{azaria2023internal,
  title={The Internal State of an {LLM} Knows When It's Lying},
  author={Azaria, Amos and Mitchell, Tom},
  booktitle={Findings of the Association for Computational Linguistics: EMNLP}, year={2023}, note={arXiv:2304.13734}
}

@inproceedings{burns2023ccs,
  title={Discovering Latent Knowledge in Language Models Without Supervision},
  author={Burns, Collin and Ye, Haotian and Klein, Dan and Steinhardt, Jacob},
  booktitle={International Conference on Learning Representations (ICLR)}, year={2023}, note={arXiv:2212.03827}
}

@inproceedings{burger2024truth,
  title={Truth is Universal: Robust Detection of Lies in {LLMs}},
  author={B{\"u}rger, Lennart and Hamprecht, Fred A. and Nadler, Boaz},
  booktitle={Advances in Neural Information Processing Systems (NeurIPS)}, year={2024}, note={arXiv:2407.12831}
}

@inproceedings{mallen2024quirky,
  title={Eliciting Latent Knowledge from Quirky Language Models},
  author={Mallen, Alex and Brumley, Madeline and Kharchenko, Julia and Belrose, Nora},
  booktitle={Conference on Language Modeling (COLM)}, year={2024}, note={arXiv:2312.01037}
}

@article{farquhar2023challenges,
  title={Challenges with Unsupervised {LLM} Knowledge Discovery},
  author={Farquhar, Sebastian and Varma, Vikrant and Kenton, Zachary and Gasteiger, Johannes and Mikulik, Vladimir and Shah, Rohin},
  journal={arXiv preprint arXiv:2312.10029}, year={2023}
}

@article{alain2016understanding,
  title={Understanding Intermediate Layers Using Linear Classifier Probes},
  author={Alain, Guillaume and Bengio, Yoshua},
  journal={arXiv preprint arXiv:1610.01644}, year={2016}
}

@article{zou2023repe,
  title={Representation Engineering: A Top-Down Approach to {AI} Transparency},
  author={Zou, Andy and Phan, Long and Chen, Sarah and Campbell, James and Guo, Phillip and Ren, Richard and Pan, Alexander and Yin, Xuwang and Mazeika, Mantas and others},
  journal={arXiv preprint arXiv:2310.01405}, year={2023}
}

@inproceedings{zou2024circuit,
  title={Improving Alignment and Robustness with Circuit Breakers},
  author={Zou, Andy and Phan, Long and Wang, Justin and Duenas, Daniel and Lin, Maxwell and Andriushchenko, Maksym and Wang, Rowan and Kolter, J. Zico and Fredrikson, Matt and Hendrycks, Dan},
  booktitle={Advances in Neural Information Processing Systems (NeurIPS)}, year={2024}, note={arXiv:2406.04313}
}

@inproceedings{li2023iti,
  title={Inference-Time Intervention: Eliciting Truthful Answers from a Language Model},
  author={Li, Kenneth and Patel, Oam and Vi{\'e}gas, Fernanda and Pfister, Hanspeter and Wattenberg, Martin},
  booktitle={Advances in Neural Information Processing Systems (NeurIPS)}, year={2023}, note={arXiv:2306.03341}
}

@article{turner2023activation,
  title={Activation Addition: Steering Language Models Without Optimization},
  author={Turner, Alexander Matt and Thiergart, Lisa and Leech, Gavin and Udell, David and Vazquez, Juan J. and Mini, Ulisse and MacDiarmid, Monte},
  journal={arXiv preprint arXiv:2308.10248}, year={2023}
}

@inproceedings{rimsky2024steering,
  title={Steering Llama 2 via Contrastive Activation Addition},
  author={Rimsky, Nina and Gabrieli, Nick and Schulz, Julian and Tong, Meg and Hubinger, Evan and Turner, Alexander Matt},
  booktitle={Annual Meeting of the Association for Computational Linguistics (ACL)}, year={2024}, note={arXiv:2312.06681}
}

@article{campbell2023localizing,
  title={Localizing Lying in Llama: Understanding Instructed Dishonesty on True-False Questions Through Prompting, Probing, and Patching},
  author={Campbell, James and Ren, Richard and Guo, Phillip},
  journal={arXiv preprint arXiv:2311.15131}, year={2023}
}

@inproceedings{goldowskydill2025deception,
  title={Detecting Strategic Deception with Linear Probes},
  author={Goldowsky-Dill, Nicholas and Chughtai, Bilal and Heimersheim, Stefan and Hobbhahn, Marius},
  booktitle={International Conference on Machine Learning (ICML)}, year={2025}, note={arXiv:2502.03407}
}

@inproceedings{pacchiardi2023liar,
  title={How to Catch an {AI} Liar: Lie Detection in Black-Box {LLMs} by Asking Unrelated Questions},
  author={Pacchiardi, Lorenzo and Chan, Alex J. and Mindermann, S{\"o}ren and Moscovitz, Ilan and Pan, Alexa Y. and Gal, Yarin and Evans, Owain and Brauner, Jan},
  booktitle={International Conference on Learning Representations (ICLR)}, year={2024}, note={arXiv:2309.15840}
}

@misc{macdiarmid2024probes,
  title={Simple Probes Can Catch Sleeper Agents},
  author={MacDiarmid, Monte and Maxwell, Timothy and Schiefer, Nicholas and others},
  year={2024}, howpublished={Anthropic Alignment Science Blog},
  note={\url{https://www.anthropic.com/research/probes-catch-sleeper-agents}}
}

@article{meinke2024scheming,
  title={Frontier Models are Capable of In-context Scheming},
  author={Meinke, Alexander and Schoen, Bronson and Scheurer, J{\'e}r{\'e}my and Balesni, Mikita and Shah, Rusheb and Hobbhahn, Marius},
  journal={arXiv preprint arXiv:2412.04984}, year={2024}
}

@article{ren2025mask,
  title={The {MASK} Benchmark: Disentangling Honesty From Accuracy in {AI} Systems},
  author={Ren, Richard and Agarwal, Arunim and Mazeika, Mantas and Menghini, Cristina and Vacareanu, Robert and Kenstler, Brad and Yang, Mick and Barrass, Isabelle and Gatti, Alice and Yin, Xuwang and Trevino, Eduardo and Geralnik, Matias and Khoja, Adam and Lee, Dean and Yue, Summer and Hendrycks, Dan},
  journal={arXiv preprint arXiv:2503.03750}, year={2025}
}

@article{hubinger2024sleeper,
  title={Sleeper Agents: Training Deceptive {LLMs} that Persist Through Safety Training},
  author={Hubinger, Evan and Denison, Carson and Mu, Jesse and others},
  journal={arXiv preprint arXiv:2401.05566}, year={2024}
}

@inproceedings{vanderweij2024sandbagging,
  title={{AI} Sandbagging: Language Models can Strategically Underperform on Evaluations},
  author={van der Weij, Teun and Hofst{\"a}tter, Felix and Jaffe, Ollie and Brown, Samuel F. and Ward, Francis Rhys},
  booktitle={International Conference on Learning Representations (ICLR)}, year={2025}, note={arXiv:2406.07358}
}

@inproceedings{greenblatt2024passwordlocked,
  title={Stress-Testing Capability Elicitation With Password-Locked Models},
  author={Greenblatt, Ryan and Roger, Fabien and Krasheninnikov, Dmitrii and Krueger, David},
  booktitle={Advances in Neural Information Processing Systems (NeurIPS)}, year={2024}, note={arXiv:2405.19550}
}

@inproceedings{hofstaetter2025elicitation,
  title={The Elicitation Game: Evaluating Capability Elicitation Techniques},
  author={Hofst{\"a}tter, Felix and van der Weij, Teun and Teoh, Jayden and Djoneva, Rada and Bartsch, Henning and Ward, Francis Rhys},
  booktitle={International Conference on Machine Learning (ICML)}, year={2025}, note={arXiv:2502.02180}
}

@inproceedings{tice2024noise,
  title={Noise Injection Reveals Hidden Capabilities of Sandbagging Language Models},
  author={Tice, Cameron and Kreer, Philipp Alexander and Helm-Burger, Nathan and Shahani, Prithviraj Singh and Ryzhenkov, Fedor and Roger, Fabien and Neo, Clement and Haimes, Jacob and Hofst{\"a}tter, Felix and van der Weij, Teun},
  booktitle={Advances in Neural Information Processing Systems (NeurIPS)}, year={2025}, note={arXiv:2412.01784}
}

@inproceedings{li2024wmdp,
  title={The {WMDP} Benchmark: Measuring and Reducing Malicious Use with Unlearning},
  author={Li, Nathaniel and Pan, Alexander and Gopal, Anjali and Yue, Summer and Berrios, Daniel and others},
  booktitle={International Conference on Machine Learning (ICML)}, year={2024}, note={arXiv:2403.03218}
}

@inproceedings{zhang2024npo,
  title={Negative Preference Optimization: From Catastrophic Collapse to Effective Unlearning},
  author={Zhang, Ruiqi and Lin, Licong and Bai, Yu and Mei, Song},
  booktitle={Conference on Language Modeling (COLM)}, year={2024}, note={arXiv:2404.05868}
}

@inproceedings{yao2023llmunlearning,
  title={Large Language Model Unlearning},
  author={Yao, Yuanshun and Xu, Xiaojun and Liu, Yang},
  booktitle={Advances in Neural Information Processing Systems (NeurIPS)}, year={2024}, note={arXiv:2310.10683}
}

@article{lucki2024adversarial,
  title={An Adversarial Perspective on Machine Unlearning for {AI} Safety},
  author={{\L}ucki, Jakub and Wei, Boyi and Huang, Yangsibo and Henderson, Peter and Tram{\`e}r, Florian and Rando, Javier},
  journal={Transactions on Machine Learning Research (TMLR)}, year={2025}, note={arXiv:2409.18025}
}

@inproceedings{hu2024jogging,
  title={Unlearning or Obfuscating? Jogging the Memory of Unlearned {LLMs} via Benign Relearning},
  author={Hu, Shengyuan and Fu, Yiwei and Wu, Zhiwei Steven and Smith, Virginia},
  booktitle={International Conference on Learning Representations (ICLR)}, year={2025}, note={arXiv:2406.13356}
}

@article{deeb2024doremove,
  title={Do Unlearning Methods Remove Information from Language Model Weights?},
  author={Deeb, Aghyad and Roger, Fabien},
  journal={arXiv preprint arXiv:2410.08827}, year={2024}
}

@article{che2025modeltampering,
  title={Model Tampering Attacks Enable More Rigorous Evaluations of {LLM} Capabilities},
  author={Che, Zora and Casper, Stephen and Kirk, Robert and Satheesh, Anirudh and Slocum, Stewart and McKinney, Lev E. and Gandikota, Rohit and Ewart, Aidan and Rosati, Domenic and Wu, Zichu and Cai, Zikui and Chughtai, Bilal and Gal, Yarin and Huang, Furong and Hadfield-Menell, Dylan},
  journal={Transactions on Machine Learning Research (TMLR)}, year={2025}, note={arXiv:2502.05209}
}

@article{lynch2024eight,
  title={Eight Methods to Evaluate Robust Unlearning in {LLMs}},
  author={Lynch, Aengus and Guo, Phillip and Ewart, Aidan and Casper, Stephen and Hadfield-Menell, Dylan},
  journal={arXiv preprint arXiv:2402.16835}, year={2024}
}

@inproceedings{bailey2025obfuscated,
  title={Obfuscated Activations Bypass {LLM} Latent-Space Defenses},
  author={Bailey, Luke and Serrano, Alex and Sheshadri, Abhay and Seleznyov, Mikhail and Taylor, Jordan and Jenner, Erik and Hilton, Jacob and Casper, Stephen and Guestrin, Carlos and Emmons, Scott},
  booktitle={International Conference on Learning Representations (ICLR)}, year={2026}, note={arXiv:2412.09565}
}

@article{lykken1959gsr,
  title={The {GSR} in the detection of guilt},
  author={Lykken, David T.},
  journal={Journal of Applied Psychology}, volume={43}, number={6}, pages={385--388}, year={1959},
  note={DOI:10.1037/h0046060}
}

@book{verschuere2011memory,
  title={Memory Detection: Theory and Application of the Concealed Information Test},
  editor={Verschuere, Bruno and Ben-Shakhar, Gershon and Meijer, Ewout},
  publisher={Cambridge University Press}, year={2011}, note={DOI:10.1017/CBO9780511975196}
}

@article{benshakhar2003validity,
  title={The validity of psychophysiological detection of information with the Guilty Knowledge Test: A meta-analytic review},
  author={Ben-Shakhar, Gershon and Elaad, Eitan},
  journal={Journal of Applied Psychology}, volume={88}, number={1}, pages={131--151}, year={2003},
  note={DOI:10.1037/0021-9010.88.1.131}
}

@article{farwell1991truth,
  title={The truth will out: Interrogative polygraphy ("lie detection") with event-related brain potentials},
  author={Farwell, Lawrence A. and Donchin, Emanuel},
  journal={Psychophysiology}, volume={28}, number={5}, pages={531--547}, year={1991},
  note={DOI:10.1111/j.1469-8986.1991.tb01990.x}
}

@misc{hopkins2026liedetectors,
  title={Fine-Tuned Lie Detectors Fail to Generalize},
  author={Hopkins, Jack and Khullar, Dipika and Wang, Rowan and Roger, Fabien},
  year={2026}, howpublished={Anthropic Alignment Science Blog},
  note={\url{https://alignment.anthropic.com/2026/lie-detectors/}}
}
\bibliographystyle{iclr2027_conference}

\appendix
\section{Standardizing activations for massive-activation models}
\label{app:norm}
Some models place a few dimensions of the residual stream at very large magnitudes, which dominate a
raw dot product and make the mass-mean readout fragile. Before forming the item-contrast direction and
the projections, we standardize each activation dimension using statistics computed over the option
spans of the fit set,
\begin{equation}
\tilde{x}_i \;=\; (x_i - \mu_i)/\sigma_i ,
\end{equation}
where $\mu_i$ and $\sigma_i$ are the mean and standard deviation of dimension $i$ over the fit-set
option spans, then compute $w^{\ell}$ and the scores on the standardized spans $\tilde{x}$. An equivalent logistic readout, trained on the standardized correct-option and
distractor spans, gives similar results. On Gemma-3 this raises recognition from $0.389$ with the raw readout
to $0.735$ with the standardized readout.

\section{Readout ablations}
\label{app:ablations}

\begin{figure}[h]
\centering
\includegraphics[width=0.9\linewidth]{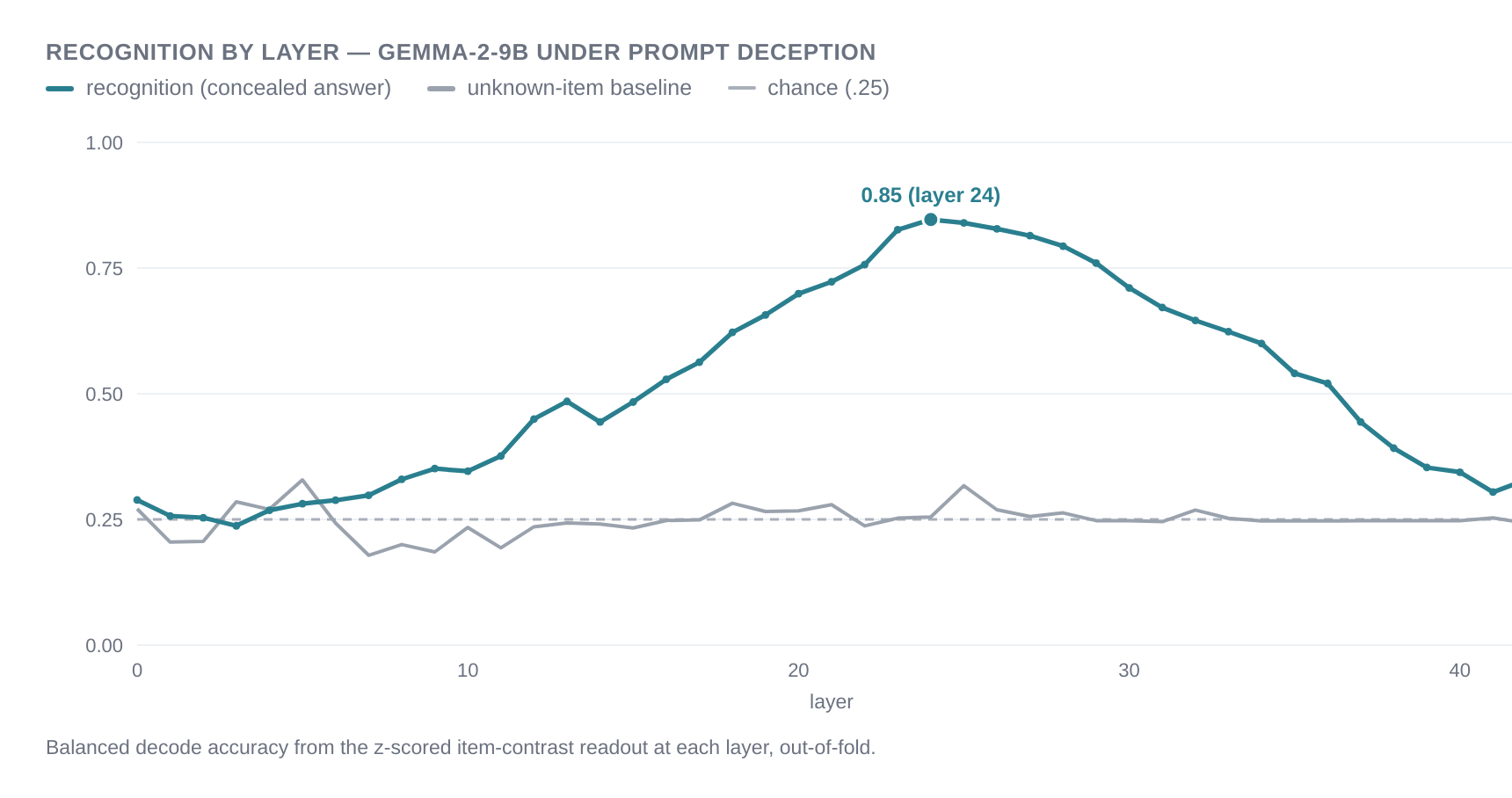}
\caption{Recognition by layer under prompt deception. The concealed answer is decodable only in the
middle layers; the unknown-item baseline stays near chance everywhere.}
\label{fig:ablation_layers}
\end{figure}

\paragraph{Where the signal lives.} We run the readout at each layer of \texttt{gemma-2-9b} under prompt
deception (Figure~\ref{fig:ablation_layers}). Recognition of the concealed answer sits at the
unknown-item baseline in the early layers, rises to $0.85$ at layer $24$ of $43$, and falls back toward
the baseline in the late layers, while the unknown-item baseline stays near chance throughout. The
signal is a middle-layer phenomenon and is weak at the layers nearest the output, which is why we select
the layer by cross-validation rather than reading a fixed depth.

\paragraph{How little supervision it needs.} Fitting the item-contrast direction at the best layer on as
few as $10$ of the model's own known questions already decodes the concealed answer at $0.82$ (mean over
five seeds), and accuracy is flat through $160$ items ($0.80$ to $0.83$). The direction is highly
data-efficient, so the readout needs only a small labeled set of known questions.

\paragraph{Where in the option to read.} PIR reads each option's last token. Comparing read positions on
\texttt{gemma-2-9b} under prompt deception, the last-token span decodes the concealed answer at $0.85$, a
mean over the option's tokens at $0.79$, and the first token at $0.53$, against an unknown-item baseline
of $0.25$ to $0.32$. The recognition signal concentrates toward the end of the option, so the last-token
choice is the strongest of the three, with mean-pooling a close second.

\section{Confound controls and statistics}
\label{app:stats}

\paragraph{Belief-versus-output test.} A within-condition probe can score well by reading the output and
negating it, since with two options the correct answer is the negation of a lie. On a binary task such a
probe reached $0.985$. We control for this in two ways. We use $k>2$ counterbalanced multiple choice, so
the correct answer is not the negation of the output. And we fit the direction on trials where the model
answers honestly and test it on trials where it conceals, so a signal that survives reads the belief. The
binary probe drops to chance under this test, while the item-contrast readout survives it.

\paragraph{Statistics.} Every headline claim is re-analyzed with a nested cross-validation point
estimate (which removes best-over-layers selection), a permutation null taken as the max over layers (so
the null pays the multiple-comparison cost), bootstrap resampling, and five fold seeds. Table~\ref{tab:stats}
gives three regimes. The prompt-deception row uses \texttt{gemma-3} with the raw readout, whose few very
large activation dimensions depress every raw estimate for this model, the honest control included, so
that row reads low across the board; the standardized readout raises \texttt{gemma-3} recognition to
$0.735$ (Appendix~\ref{app:norm}), and we keep the raw nested-CV values here only to exercise the
permutation and selection controls. The boundary holds. RMU is the decisive row. The readout works (control $0.751$), yet
recognition sits at the unknown-item baseline (nested-CV $0.300$ against a baseline of $0.339$, the nested-CV estimate
of the $0.39$ reported in the main text, permutation $p=0.086$), so the drop reflects a genuine absence
while the readout still works. A two-one-sided-tests check places the RMU recognition-minus-baseline gap
within a small margin of zero.

\begin{table}[h]
\centering
\caption{Nested-CV recognition by regime (raw readout), with a permutation-null p-value and a positive
control.}
\label{tab:stats}
\begin{tabular}{lcccc}
\toprule
regime & recognition & baseline & perm-$p$ & control \\
\midrule
prompt deception (gemma-3, MMLU) & 0.420 & 0.200 & $<$0.001 & 0.340 \\
output-gating (sandbag, WMDP) & 0.891 & 0.360 & $<$0.001 & 0.881 \\
representation erasure (RMU, WMDP) & 0.300 & 0.339 & 0.086 & 0.751 \\
\bottomrule
\end{tabular}
\end{table}

\paragraph{A label-free direction.} The item-contrast direction can be found with no labels, as the top
principal component of the per-question centered option spans (labels are used only to pick the component
by cross-validation and to evaluate). It nearly matches the supervised direction, with MMLU $0.812$ against
$0.825$, ARC $0.889$ against $0.898$, and WMDP-cyber $0.788$ against $0.854$ (chance $0.25$).

\paragraph{Cross-dataset transfer.} Fitting the direction on one dataset and applying it to another
without refitting is near-lossless. Every off-diagonal is within $0.01$ to $0.05$ of the target's own
within-dataset accuracy and far above the unknown-item baseline. A WMDP-fit direction reads MMLU at $0.815$ and
ARC at $0.900$, and an MMLU-fit direction reads WMDP at $0.803$. The recognition direction is a single
model-level correct-option direction that transfers across datasets.

\section{Confident-concealment frontier}
\label{app:frontier}
We test directly the claim that internals dominate black-box detection when a model conceals confidently.
On \texttt{gemma-2-9b} under prompt deception, we compare a concealment run to an honest run on the same
known questions, so the black-box negative is a fixed confident reference rather than the variable
unknown set. We then bin the concealed questions by the model's output confidence on its emitted wrong
answer. A black-box detector (logistic regression on output-distribution shape, entropy, max-probability,
and margin) and the internal divergence detector each separate concealment from honest.

\begin{figure}[h]
\centering
\includegraphics[width=0.72\linewidth]{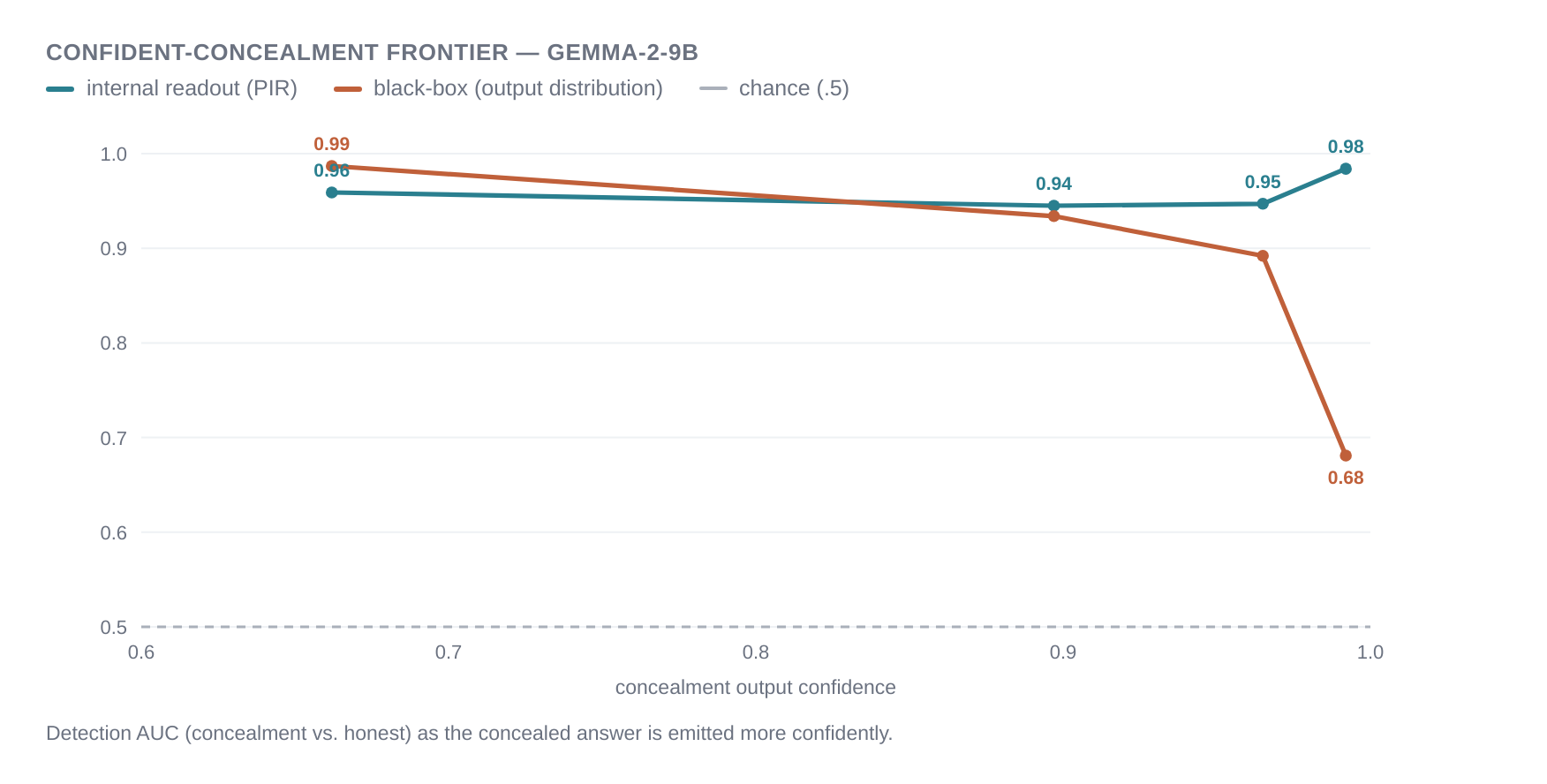}
\caption{Confident-concealment frontier. As the concealed answer is emitted more confidently, black-box
detection falls toward chance while the internal readout holds.}
\label{fig:frontier}
\end{figure}

As concealment becomes more confident, the concealed output looks like an honest answer and the black-box
AUC falls from $0.99$ to $0.68$, while the internal readout stays near $0.95$ across the range
(Figure~\ref{fig:frontier}). The internal signal reads the recognized answer from activations, so it does
not depend on how confidently the model speaks. This is one organism on one model, and the black-box
detector degrades but does not reach chance.

\section{The adversarial anti-probe organism}
\label{app:adversarial}
We train a LoRA organism with three objectives at once. It keeps the capability under the trigger,
sandbags without it, and drives the item-contrast recognition margin to zero on the untriggered spans. We
sweep the weight $\lambda$ on the evasion term (Table~\ref{tab:antiprobe}).

\begin{table}[h]
\centering
\caption{Anti-probe dose sweep. Recognition is on the untriggered known spans against a fresh
question-disjoint probe.}
\label{tab:antiprobe}
\begin{tabular}{lccc}
\toprule
$\lambda$ & triggered capability & recognition (untriggered) & baseline \\
\midrule
0.5 & 0.736 & 0.77 & 0.38 \\
1.0 & 0.714 & 0.68 & 0.31 \\
2.0 & 0.714 & 0.34 & 0.31 \\
3.0 & 0.24 (collapsed) & -- & -- \\
\bottomrule
\end{tabular}
\end{table}

At $\lambda=2$ the model keeps its capability (triggered $0.714$, triggered spans still readable at
$0.83$) and drives the untriggered recognition to the unknown-item baseline. We verified this under
question-disjoint cross-validation, which corrected an earlier in-sample result. A fixed honest-calibrated
monitor read $0.681$ in-sample but $0.367$ under proper cross-validation, at the unknown-item baseline, and an
adversary trained directly against that fixed direction reached $0.407$. The in-sample inflation appears
only in this collapsed-span regime, where the option spans carry a large per-question self-correlation
(cosine $0.997$).

The attack removes the signal from the locus the monitor reads. It does not merely move the signal off
one direction. On the untriggered
known spans a stronger linear probe, an unsupervised principal-component direction, and a nonlinear
multilayer probe all sit at the unknown-item baseline ($0.30$ to $0.35$), while all read the triggered spans at
$0.79$ to $0.85$. Below-chance untriggered accuracy ($0.11 < 0.25$) shows the model still computes the
answer somewhere. The collapse is a plausible tamper signal, but a detector for it needs a base-model
anomaly reference and an anomaly-aware adversary, which we leave open.

\end{document}